\documentclass[conference]{IEEEtran}

\let\labelindent\relax

\usepackage{times}
\usepackage{graphics,graphicx,float,booktabs,xcolor,multirow,array,color,ifthen,tabu,colortbl,dblfloatfix,url,xparse,mathtools,patchcmd,algorithm,algorithmic,amssymb,xspace,nicefrac,microtype,amsmath,amsthm,amsfonts,bm,ragged2e,tikz,stackengine,etoolbox,xpatch,enumerate,xstring,setspace,makecell,changepage,cuted,titlesec,enumitem,wrapfig,tcolorbox, wrapfig}
\usepackage[numbers]{natbib}
\usepackage{multicol, booktabs}
\usepackage[bookmarks=true]{hyperref}
\usepackage{caption}

\begin{document}

\newcommand{\our}{FACTR\xspace}
\newcommand{\yulong}[1]{\textcolor{blue}{Yulong: #1}}
\newcommand{\jason}[1]{\textcolor{green}{Jason: #1}}
\newcommand{\todo}[1]{\textcolor{red}{Todo: #1}}

\title{\textbf{FACTR}: \textbf{F}orce-\textbf{A}ttending \textbf{C}urriculum \textbf{Tr}aining for Contact-Rich Policy Learning}

\author{\vspace{2mm}
    Jason Jingzhou Liu\IEEEauthorrefmark{1}, Yulong Li\IEEEauthorrefmark{1}, Kenneth Shaw, Tony Tao, Ruslan Salakhutdinov, Deepak Pathak \\\vspace{.5mm}
    Carnegie Mellon University \\
    \IEEEauthorrefmark{1}Equal contribution
}

\makeatletter
\let\@oldmaketitle\@maketitle
\renewcommand{\@maketitle}{\@oldmaketitle
    \vspace{-0mm}
    \includegraphics[width=0.92\linewidth]{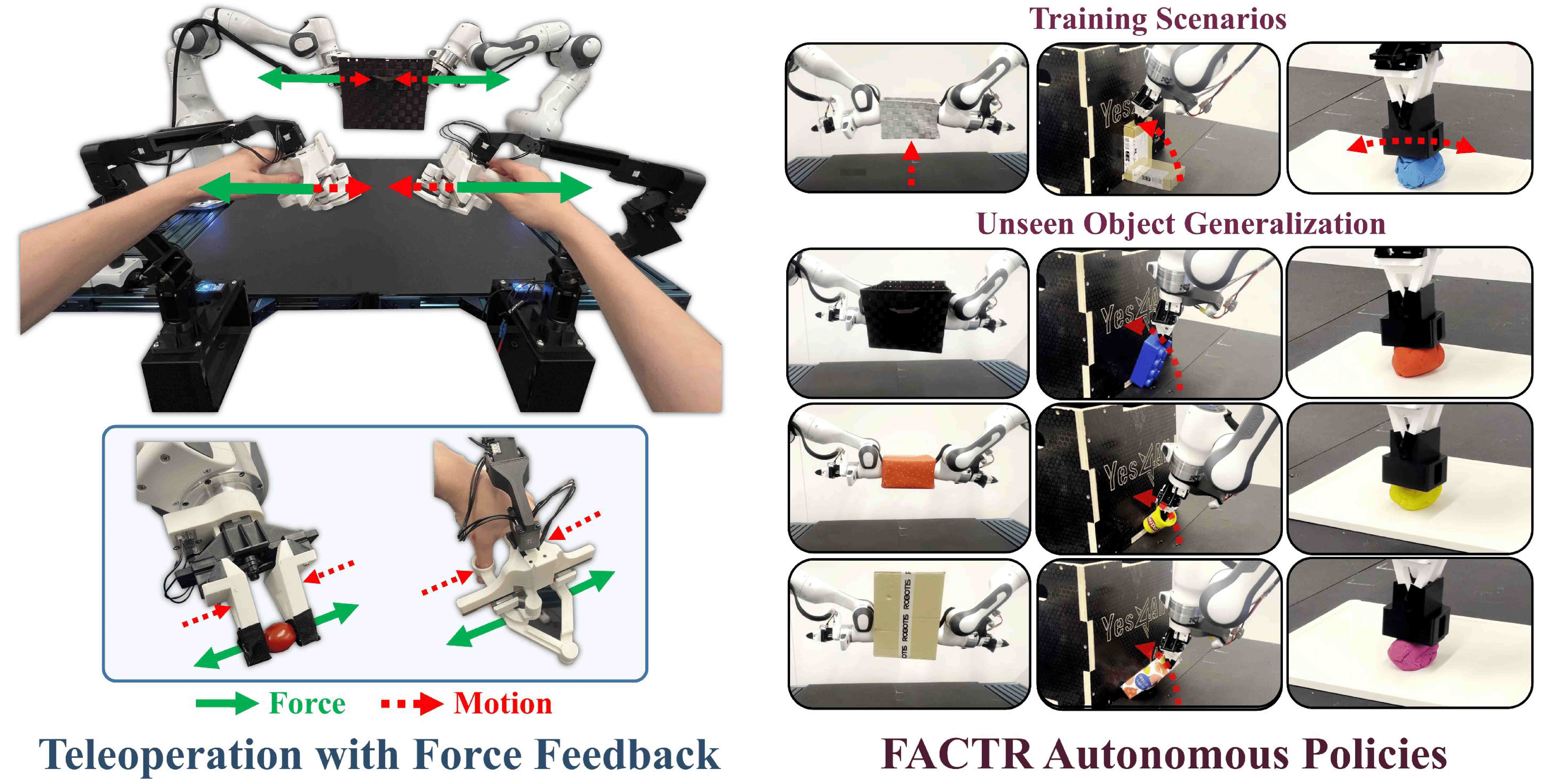}
    \centering
    \vspace{-1mm}
    \captionof{figure}{\small \textbf{FACTR.} We present \textit{Force-Attending Curriculum Training} (FACTR) -- a system that leverages robot external joint torques for both teleoperation and improving policy generalization for complex contact-rich tasks. [Left] Our low-cost leader-follower setup employs actuated servo motors to enable force feedback in both the leader arm and gripper, improving teleoperation success rate, completion time, and ease of use. [Right] FACTR's behavior cloning policy utilizes robot force information to enhance performance and generalization across objects with diverse geometries and textures in contact-rich tasks. Video results, codebases, and instructions at \url{https://jasonjzliu.com/factr/}}
    \vspace{-3mm}
    \label{fig:teaser}
}
\makeatother
\maketitle
\addtocounter{figure}{-1}

\begin{abstract}
Many contact-rich tasks humans perform, such as box pickup or rolling dough, rely on force feedback for reliable execution. However, this force information, which is readily available in most robot arms, is not commonly used in teleoperation and policy learning. Consequently, robot behavior is often limited to quasi-static kinematic tasks that do not require intricate force-feedback. In this paper, we first present a low-cost, intuitive, bilateral teleoperation setup that relays external forces of the follower arm back to the teacher arm, facilitating data collection for complex, contact-rich tasks. We then introduce FACTR, a policy learning method that employs a curriculum which corrupts the visual input with decreasing intensity throughout training. The curriculum prevents our transformer-based policy from over-fitting to the visual input and guides the policy to properly attend to the force modality. We demonstrate that by fully utilizing the force information, our method significantly improves generalization to unseen objects by 43\% compared to baseline approaches without a curriculum. Video results, codebases, and instructions at \url{https://jasonjzliu.com/factr/}
\end{abstract}

\enlargethispage{1\baselineskip} 

\section{Introduction}
Contact-rich tasks are an integral part of daily life, from lifting a box and rolling dough to cracking an egg or opening a door. These tasks, while seemingly simple, involve a complex interplay of forces and require precise adjustments based on force feedback. Humans rely heavily on this force feedback to generalize across tasks and objects, adapting seamlessly to variations in visual appearances and geometries. However, in robot learning, force information remains underutilized, even though it is readily available on many modern robotic arms, such as the Franka Panda and the KUKA LBR iiwa. Instead, most data-driven methods, including those using Behavior Cloning (BC), focus primarily on visual feedback for both data collection and policy learning, overlooking the critical role of force. This limited use of force information hinders the vision-only policies' ability to generalize to novel objects. For instance, in tasks like lifting up a box with two arms, the primary factor influencing the action is the object's geometry, while attributes such as color or texture are irrelevant. In such cases, force feedback provides a clear signal for mode switching, such as detecting when contact is established, which can facilitate object generalization compared to relying solely on vision.

\enlargethispage{1\baselineskip} 
One of the main reasons for the under-utilization of force feedback in robot learning is the lack of an intuitive and \textit{low-cost} teleoperation system that can capture force feedback during data collection itself.
Recently, low-cost leader-follower systems have become popular for teleoperation, offering intuitive control of robot arms by mirroring the joint movements of the leader arm controlled by a teleoperator to the follower arm \cite{aloha, wu2024gello}. However, these systems are typically passive (leader arm joints are not actuated) and unilateral (the leader arm does not receive information from the follower arm). This makes teleoperation difficult for dynamic, contact-rich tasks where precise force adjustments are necessary~\cite{pagliara2024safe}. To overcome this limitation, we present a bilateral low-cost teleoperation system that provides force feedback by actuating motors in the leader arm joints based on external joint torques transmitted from the follower arm (Fig.~\ref{fig:teaser} Left). By actuating the motors, we also provide active gravity compensation and resolve the kinematic redundancy due to the redundant degrees of freedom of the arm. These enhancements improve the teleoperation experience, leading to a 64.7\% increase in the task completion rate, a 37.4\% reduction in completion time, and an 83.3\% improvement in subjective ease of use across four evaluated contact-rich tasks.

The second challenge lies in effectively incorporating robot force information into policy learning. Although recent methods such as diffusion policy \cite{chi2023diffusion} and action chunking transformers \cite{aloha} achieve impressive results for fine-grained manipulation, they often fail to generalize to unseen objects with variations in object visual appearances and geometries. Humans, on the other hand, can disregard irrelevant visual details once contact is established and rely solely on force feedback to perform tasks such as lifting a box or rolling dough. Therefore, to improve generalization, we seek to incorporate force input into autonomous robot policies. However, making effective use of force information in policy learning is challenging, as policies often overfit to using visual modality~\cite{wang2020makes}, effectively disregarding force data. This issue arises because contact force signals are typically less discriminative, often remaining near zero for extended periods when the arm is not in contact with the environment during an episode. Hence, without proper care during training, policies tend to ignore force input and rely primarily on visual information. We empirically analyze this effect in Sec.~\ref{sec:analysis} and Fig.~\ref{fig:test_analysis}.

To mitigate this imbalance, we propose Force-Attending Curriculum Training (FACTR), a curriculum training strategy designed to improve the policy’s ability to effectively leverage force information. FACTR systematically reduces the reliance on visual information during training by applying operators such as Gaussian blurring or downsampling with varying scales to visual inputs. A scheduler gradually decreases the blurring scale and increases the fidelity of the visual inputs. Intuitively, this approach encourages the policy to focus more on force input during initial training phases and gradually balances force with visual inputs as training progresses. We ground this intuition with a theoretical analysis on a simplified scenario through the framework of Neural Tangent Kernels~\cite{jacot2018neural}. We explore FACTR in both the pixel space and latent space, testing various operators and scheduling strategies. Our experiments show that FACTR improves the success rate for unseen objects by an average of 40.0\% in four challenging contact-rich tasks (Fig.~\ref{fig:teaser} Right) -- box lifting, prehensile pivoting, fruit pick-and-place, and  dough rolling -- showcasing the efficacy of our force-attending curriculum training.

\begin{figure}[t]
    \centering
    \includegraphics[width=\linewidth]{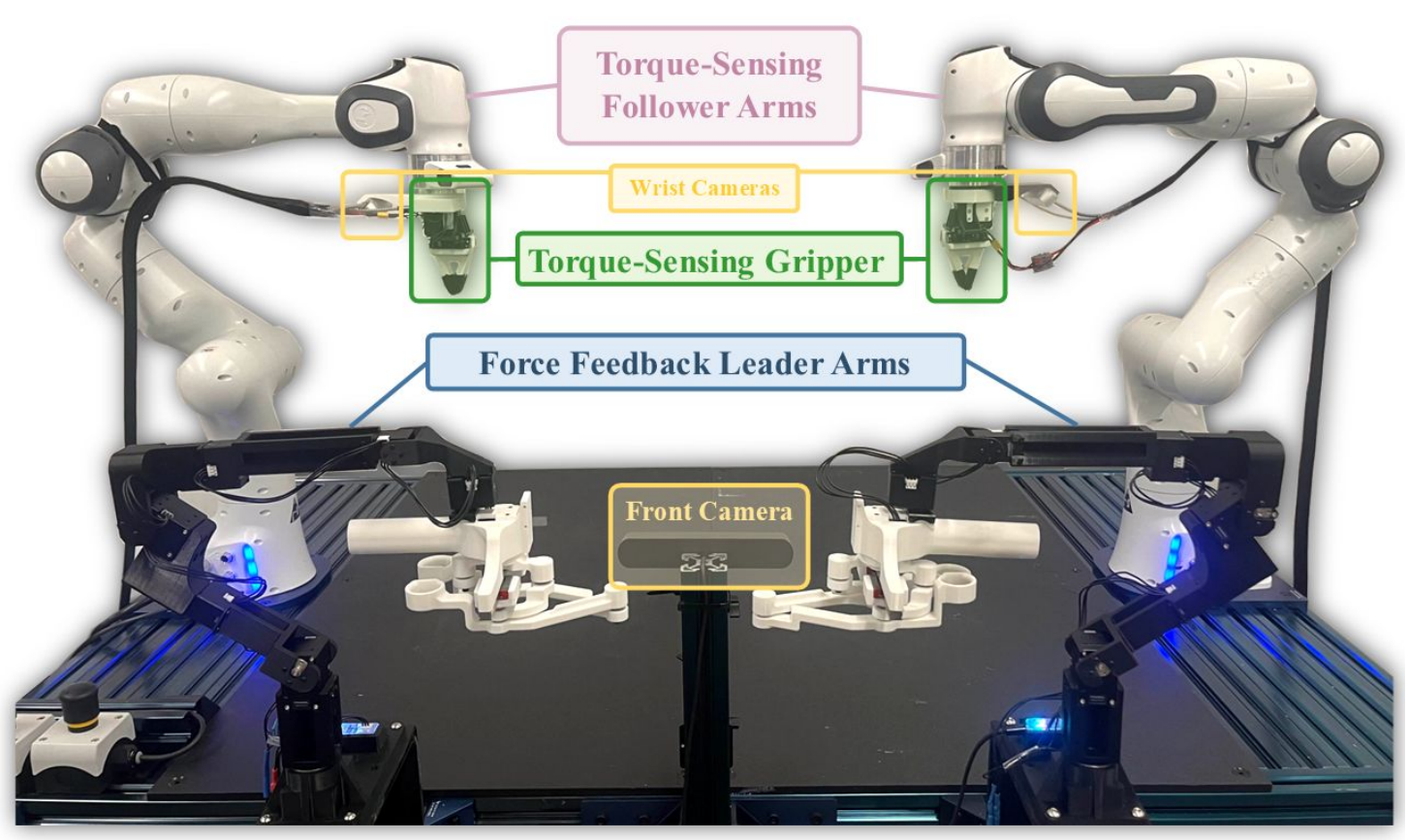}
    \caption{\small\textbf{Our low-cost bimanual teleoperation system with force-feedback.} The system features two actuated leader arms, two follower arms with external joint torque sensors (such as the Franka Panda and the KUKA LBR iiwa), a front camera and two wrist cameras.}
    \label{fig:setup}
    \vspace{-6mm}
\end{figure}

In summary, our contributions are as follows.
\begin{itemize} 
\item \textbf{Low-Cost Teleoperation with Force Feedback:} We design a low-cost bilateral leader-follower teleoperation system with force feedback, gravity compensation, and redundancy resolution, demonstrating a 64.7\% improvement in task completion rate and an 83.3\% enhancement in ease of use for teleoperation through a user study.
\item \textbf{Force-Attending Curriculum Training:} We propose FACTR, a curriculum training framework that better learns to use force feedback in policy learning and achieves better generalization capability to object visual appearances and geometries. Evaluated on four challenging contact-rich tasks, FACTR improves the performance of autonomous policies by 40.0\% compared to policies with force information as part of input but trained without FACTR.
\end{itemize}

\section{Related Works}

\subsection{Imitation Learning with Force}

Imitation learning has recently experienced significant advancements, driven by the development of more effective algorithms that leverage demonstrations to train robotic policies~\cite{mandlekar2021matters, chi2023diffusion}. Although traditional approaches primarily rely on visual and joint position inputs, many real-world tasks require explicit force feedback to improve stability, adaptability, and safety~\cite{mason1981compliance, hogan1984impedance}. 
Recent work has applied learning methods to train with demonstrations that incorporate gripper force or tactile signals, resulting in policies capable of handling small, fragile objects and performing contact-intensive tasks such as vegetable peeling~\cite{liu2024forcemimic, xie2024just, li2024haptic}. However, utilizing force data from the robot arms, such as joint torques, remains underexplored. One approach involves using an end-effector force sensor to estimate compliance parameters or virtual position targets through kinesthetic teaching and force tensors~\cite{hou2024adaptive, chen2025dexforce}. Another method infers a 6D wrench for low-level control by integrating torque sensing into a diffusion policy~\cite{wu2024tacdiffusion}.

However, naively incorporating force feedback into policy learning can lead to overfitting to visual information, causing the policy to disregard force input. FoAR~\cite{he2024foar} explicitly predicts contact and non-contact phases to regulate the fusion of vision and force modalities, which requires additional data labeling. We propose FACTR to effectively incorporate force and vision input into policy through a curriculum, enabling policies to leverage force for improved object generalization.

\subsection{Low-Cost Teleoperation Systems with Force Feedback}

Parallel to advances in imitation learning, significant efforts have been made to collect low-cost and high-quality data with hand-held grippers~\cite{song2020grasping, umi} or leader-follower systems~\cite{aloha, wu2024gello, bidex}. Hand-held grippers naturally provide force feedback to the operator, but they do not directly record force data. Recent work has added force sensors to hand-held grippers to address this limitation~\cite{liu2024forcemimic}. However, hand-held grippers are in general limited by the kinematic differences between humans and robots, resulting in commands that might be unachievable for the robots.
Although the leader-follower systems are not prone to this limitation, they often lack force feedback, impairing their effectiveness in contact-rich tasks. Recently, Kobayashi et al.~\cite{kobayashi2024alpha} implemented a bilateral leader-follower teleoperation system where in addition to the follower following the joint positions of the leader, the leader also gets an additional torque if there is a difference in its joint position from that of the follower.
However, when the follower arm is in motion without contact, this system causes the operator to experience inertial, frictional, and other dynamic forces of the follower, reducing the ease of use and precision of the system~\cite{siciliano2008springer}. Our approach introduces an alternative bilateral teleoperation method by relaying only external joint torques from the follower arm back to the leader arm, providing force feedback without impairing operational precision.

\section{FACTR Low-Cost Bilateral Teleoperation}
Leader-follower systems, such as GELLO~\cite{wu2024gello} or ALOHA~\cite{aloha}, offer a simple and cost-effective solution to teleoperation in manipulation tasks. These systems feature kinematically equivalent leader and follower arms, allowing intuitive control through joint space mapping, where the leader's joint positions are mirrored as targets for the follower. This setup lets users naturally feel the follower arms' kinematic constraints. However, most implementations lack force feedback, preventing users from sensing the geometric constraints of the environment, which is crucial for teleoperating contact-rich tasks~\cite{pagliara2024safe}. Instead, those leader arms are mostly passive, lacking active motor torque actuation, despite being equipped with servo motors capable of actuation. Furthermore, the lack of active torque means the leader arms require external structural frames and rubber bands or strings to achieve gravity compensation, reducing portability~\cite{aloha}.

In this paper, we aim to fully leverage the servo motors in the leader arm and gripper to achieve force-feedback enabled teleoperation with affordable hardware. Similarly to GELLO~\cite{wu2024gello}, our leader arms use off-the-shelf servos and 3D-printed components, forming a scaled-down but kinematically equivalent version of the follower arms, as shown in Fig.~\ref{fig:setup}. By actuating the servo motors, we introduce force feedback, customizable redundancy resolution through nullspace projection, gravity and friction compensation, and joint limit avoidance. These functionalities augment the teleoperation experience while still using low-cost hardware to provide functions that are usually only available with much more expensive teleoperation devices. Please see Appendix~\ref{sec:cost} for a detailed Bill of Materials. 

\subsection{Force Feedback}
Force feedback provides the operator with a tangible sense of interaction with the environment, allowing more intuitive and delicate manipulation, especially in contact-rich tasks or tasks with limited visual feedback \cite{pagliara2024safe}. We implement a control law that relays external joint torques sensed by the follower arm to the leader arm, allowing the operator to feel the physical constraints experienced by the follower arm:
\begin{equation}
    \mathbf{\tau}_{feedback} = \mu_f \mathbf{K}_{f,p} \mathbf{\tau}_{ext} - \mathbf{K}_{f,d} \mathbf{\dot{q}}
\end{equation}
where $\mu_f$ is a scalar constant, $\mathbf{\tau}_{ext}$ is the external joint torque sensed by the follower arm, $\mathbf{K}_{f,p}$ and $\mathbf{K}_{f,d}$ are the PD gains for the force feedback, respectively. Here, $\mathbf{K}_{f,p}$ is calculated as the ratio between the maximum torque of the leader and that of the follower, and $\mathbf{K}_{f,d} \mathbf{\dot{q}}$ helps reduce oscillations in the leader arm when the follower arm is in contact. We note that $\mathbf{\tau}_{ext}$ is a readily available measurement in various collaborative robot manipulators, such as the Franka Panda and the KUKA LBR iiwa. In particular, we implement mediated force feedback by scaling down $\mathbf{\tau}_{ext}$ with $\mu_f$, which has been shown to improve the accuracy of the operation while reducing the cognitive load of the operator~\cite{ZHU2021103674}. Furthermore, we highlight that our implementation only transmits external forces from the follower to the leader; as a result, the operator does not experience the internal friction and inertia of the follower arm during motion, providing a clearer perception of the environment~\cite{siciliano2008springer}.

In addition, we implement force feedback for the parallel-jaw gripper. Since our servo-based gripper does not contain an external force sensor, we utilize the present current reading of the gripper servo to provide force feedback as follows:
\begin{equation}
    \tau_{h, t} = \alpha (-k_h I_{g, t}) + (1 - \alpha)\tau_{h, t-1}
\end{equation}
where $\tau_{h, t}$ is the force feedback torque sent to the gripper leader device, $I_{g, t}$ is the present current reading from the follower gripper, and $\alpha$ is the smoothing factor for the EMA filter. Our system sets $\alpha=0.1$ which provides a good user experience.

\subsection{Customizable Redundancy Resolution}
\begin{figure}
    \centering
    \includegraphics[width=\linewidth]{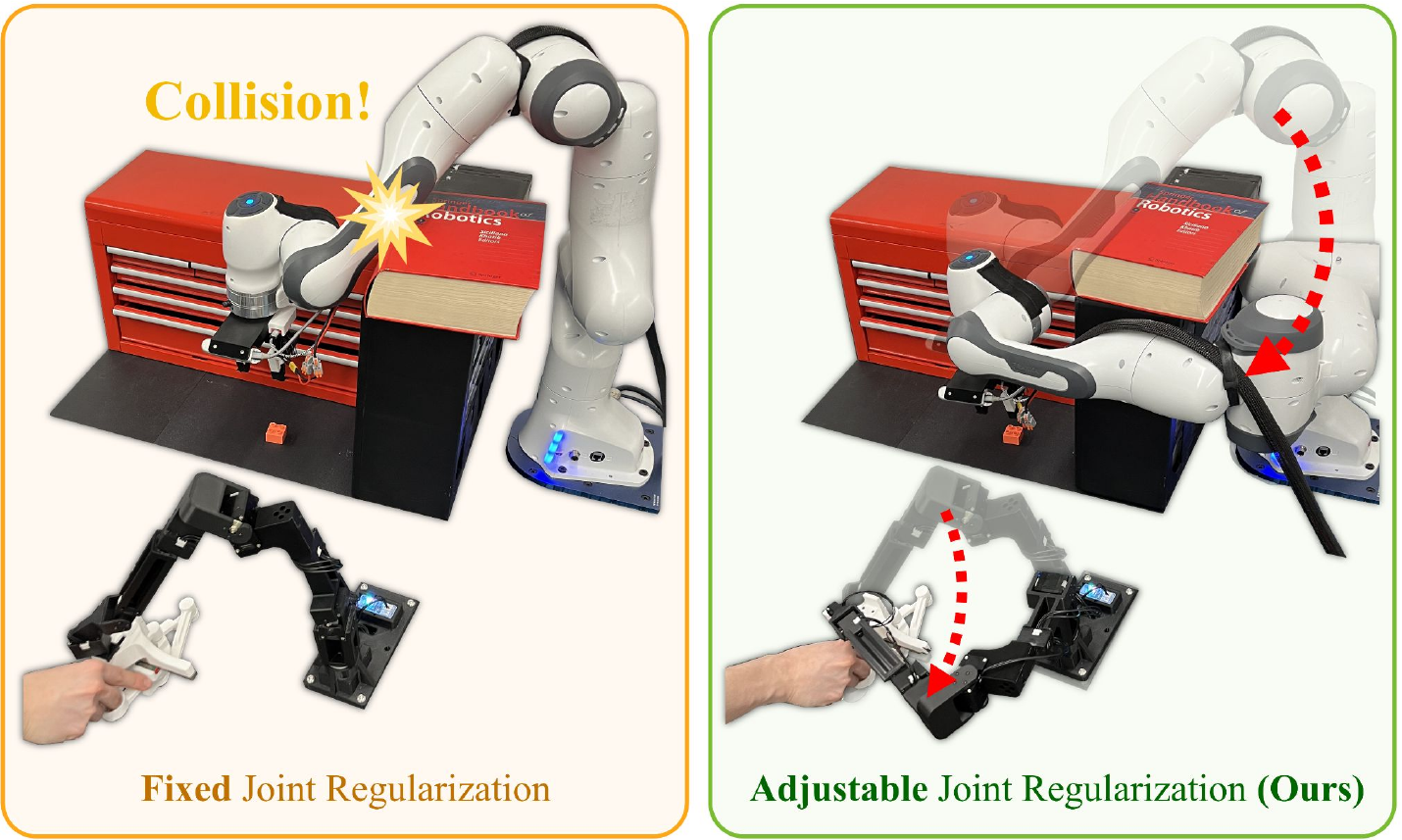}
    \caption{\small \textbf{Customizable Joint Regularization} [Left] Without the flexibility to define the resting joint configuration $q_{rest}$, the arm’s reachability is restricted, leading to collisions in confined spaces. [Right] Our leader arm allows the user to define custom resting joint $q_{rest}$, helping the follower arm reach targets in confined spaces.
    }
    \vspace{-6mm}
    \label{fig:redundancy_resolution}
\end{figure}

For kinematic redundant manipulators, without regulating the joint space, the manipulator tends to drift into undesirable configurations under the influence of gravity during teleoperation. Approaches like Gello~\cite{wu2024gello} rely on mechanical components, such as springs, to regularize the joint space. However, these components introduce non-uniform, configuration-dependent wrenches at the end-effector, resulting in an unintuitive teleoperation experience. In addition, using mechanical joint regularization effectively prevents the user from setting custom joint regularization targets for redundancy resolution. In confined-space manipulation settings, the inability to control the joint regularization target can impair the arm's reachability, as demonstrated in Fig.~\ref{fig:redundancy_resolution}. 

In contrast, our proposed method leverages the following null-space projection control law to regulate joint positions \cite{khatib1987}, which stabilizes the joint-space at any user-defined desirable posture without imposing additional end-effector wrenches, regardless of the arm’s configuration:
\begin{equation}
    \tau_{null} = \left(\mathbf{I} - \mathbf{J}^{\dagger} \mathbf{J}\right)
    \left(
        -\mathbf{K}_{n,p} \left(\mathbf{q}-\mathbf{q}_{rest}\right)
        - \mathbf{K}_{n,d}\mathbf{\dot{q}}
    \right)
\end{equation}
where $\mathbf{J}$ is the manipulator Jacobian matrix, $\mathbf{q}_{rest}$ is a user-defined resting posture configuration, $\mathbf{K}_{n,p}$ and $\mathbf{K}_{n,d}$ are the PD gains for the null space projection. Note that $\left(\mathbf{I} - \mathbf{J}^{\dagger} \mathbf{J}\right)$ is the null-space projector.

\subsection{Gravity Compensation}

To ensure the leader arms remain stationary, allowing the user to easily pause teleoperation, we implement gravity compensation. This is achieved by modeling the dynamics of the leader arm and computing the joint torques required to counteract dynamic forces using the recursive Newton-Euler algorithm (RNEA) for real-time inverse dynamics \cite{kevin_lynch}.
\begin{equation}
\mathbf{\tau}_{grav} = \mathbf{M}(\mathbf{q}) \mathbf{\ddot{q}} + \mathbf{C}(\mathbf{q}, \mathbf{\dot{q}}) \mathbf{\dot{q}} + \mathbf{g}(\mathbf{q}) = \text{RNEA}(\mathbf{q} , \mathbf{\dot{q}}, \mathbf{\ddot{q}} )
\end{equation}

where $\mathbf{M}(\mathbf{q})$ is the mass (or inertia) matrix, $\mathbf{C}(\mathbf{q}, \mathbf{\dot{q}})$ is the Coriolis and centrifugal matrix, and $\mathbf{g}(\mathbf{q})$ is the gravity vector.

\subsection{Additional Compensation and Controls}
To reduce the perceived friction in the leader arm during teleoperation, our system provides friction compensation $\tau_{friction}$. Furthermore, since the leader arm joints lack mechanical joint limits, we implement an artificial potential based control law to prevent users from exceeding joint limits of the follower arm in order to respect the workspace of the follower arm.  Finally, for bi-manual follower arms, the system uses Riemannian Motion Policies~\cite{RMP} for dynamic obstacle avoidance between the two follower arms. Please refer to Appendix~\ref{sec:add_control_law} for more details.

\subsection{Overall Control Law for the Leader Arm}

In summary, the control torques are defined as follows:
\begin{itemize}
    \item $\mathbf{\tau}_{feedback}$ relays external forces from the follower arm back to the leader arm, allowing the operator to sense the geometric constraints of the environment.
    \item $\mathbf{\tau}_{null}$ resolves kinematic redundancy by regulating the joints at a user-defined rest posture in the null-space.
    \item $\mathbf{\tau}_{grav}$ provides gravity compensation for the leader arm.
    \item $\mathbf{\tau}_{friction}$ compensates for the leader arm joint frictions to enable smoother teleoperation.
    \item $\mathbf{\tau}_{limit}$ prevents the joints of the leader arm from violating the joint position limits of the follower arm.
\end{itemize}

The resulting combined torque applied to the servo motors of the leader arm is defined as follows:
\begin{equation}
    \mathbf{\tau} = \mathbf{\tau}_{feedback} + \mathbf{\tau}_{null} + \mathbf{\tau}_{grav} + \mathbf{\tau}_{friction} + \mathbf{\tau}_{limit}
\end{equation}

\begin{figure*}[t]
    \centering
    \includegraphics[width=0.95\linewidth]{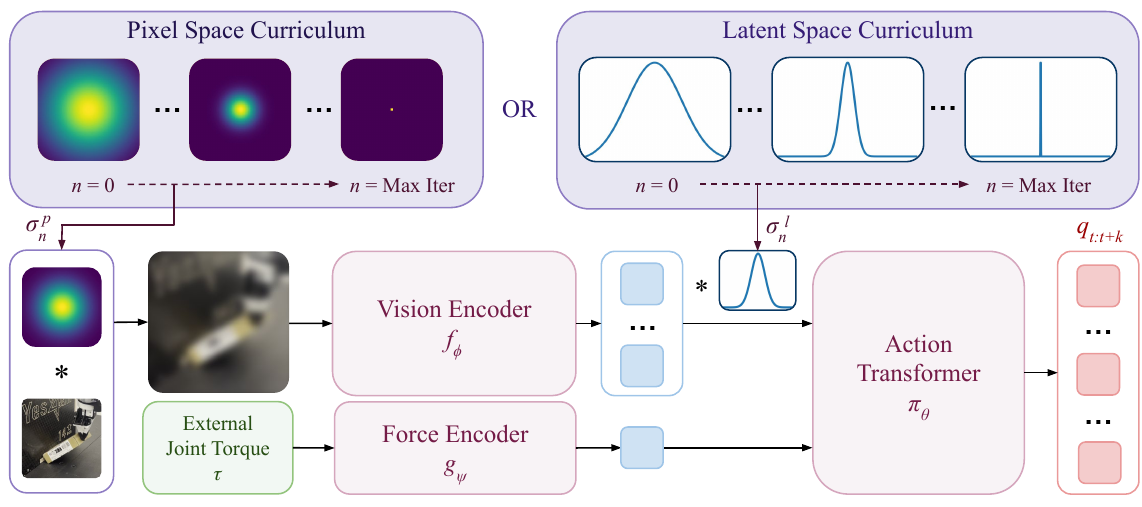}
    \caption{\small\textbf{\our} allows our policy to better integrate force information without overfitting to visual information, resulting in better generalization to objects with unseen visual appearances and geometries. Our policy takes as inputs RGB images \(I\) and external joint torque \(\tau\), which are then tokenized by a vision encoder and a force encoder before fed into an action transformer to regress joint position targets \(q_{t:t+k}\). FACTR applies a blurring operator of scale \(\sigma_n\) in either pixel or latent space, initialized at a large value then gradually decreased through the training.}
    \label{fig:method}
    \vspace{-5mm}
\end{figure*}

\section{FACTR: Force-Attending Curriculum Training}
\label{sec:method}
Naively incorporating robot force data into policy learning does not necessarily ensure policy improvement. Contact force signals often provide limited discriminative information for the policy,  as it remains close to zero for significant periods when the arm is not interacting with the environment during an episode. As a result, the policy tends to disregard force input and rely predominantly on visual information, as empirically analyzed in Sec.~\ref{sec:analysis} and Fig.~\ref{fig:test_analysis}.
 
To fully leverage the robot force data collected from our teleoperation system, we introduce Force-Attending Curriculum Training (FACTR), a training strategy designed to effectively integrate force information into policy learning. FACTR applies operators like Gaussian blur or downsampling to corrupt visual information, where the amount of visual corruption decreases throughout training. The curriculum intuitively encourages contribution from the force modality at the start of training. We ground this intuition with a theoretical analysis on a simplified scenario through the framework of Neural Tangent Kernels~\cite{jacot2018neural}.

In this section, we first present the base policy model used for learning from teleoperated demonstrations, and then motivate and describe FACTR, our curriculum training approach. Our overall method is summarized in Algorithm~\ref{alg:curriculum} and Fig.~\ref{fig:method}.

\subsection{Problem Statement and Base Model}

We consider a policy \(\pi_\theta(\cdot \mid \cdot)\) that produces a chunk of future actions of length k \(\hat{q}_{t:t+k}\) (joint positions) given (i) a visual observation \(I_t\) (image at time \(t\)), and (ii) an external joint torque reading \(\tau_t\). Our goal is to learn \(\pi_\theta\) via behavior cloning (BC) from a dataset of expert trajectories \(\mathcal{D}\). Each trajectory in \(\mathcal{D}\) comprises tuples \(\bigl(I_t, \tau_t, q_t\bigr)\), where \(q_t\) is the ground-truth (expert) joint position target at time \(t\). We let \(\hat{q}_{t:t+k}\) be the \emph{predicted} future joint position targets over the next \(k\) time steps. The loss is defined by:
\begin{equation}
\label{eq:bcloss}
\mathcal{L} \;=\; \text{MSE}\bigl(\hat{q}_{t:t+k},\, q_{t:t+k}\bigr),
\end{equation}
where \(q_{t:t+k}\) are the expert’s future joint position targets and \(\hat{q}_{t:t+k}\) are the policy’s predictions.

Our policy \(\pi_\theta\) is based on an encoder-decoder transformer that integrates vision and force modalities. Visual observations and force readings are converted into tokens, fed to the \emph{encoder}, then decoded into action tokens through cross attention.

A pre-trained vision transformer (ViT)~\cite{vit, dasari2023unbiased} is used to encode an input image \(I_t\) into a sequence of \emph{vision tokens} \(\mathbf{z}_t^V \in \mathbb{R}^{M_v \times d}\) for some number of tokens \(M_v\) and embedding dimension \(d\). An MLP-based force encoder is applied to the joint torque \(\tau_t\), resulting in a single \emph{force token}: \(\mathbf{z}_t^F \;\in\; \mathbb{R}^{1 \times d}.\)
The tokens are concatenated to form the model input:
\[
\mathbf{X}_t \;=\; \bigl[\mathbf{z}_t^V; \,\mathbf{z}_t^F\bigr] \;\in\; \mathbb{R}^{(M_v+1)\times d}.
\]

Then, a transformer encoder \(\mathrm{Enc}\) processes \(\mathbf{X}_t\) via multiple self-attention and feed-forward layers:
\[
\mathbf{H}_t^E \;=\; \mathrm{Enc}\bigl(\mathbf{X}_t\bigr) \;\in\; \mathbb{R}^{(M_v+1)\times d}.
\]
This yields the \emph{encoded} vision and force tokens.

For the decoder, we introduce \(k\) \emph{action tokens}, \(\mathbf{A} \in \mathbb{R}^{k \times d}\). A transformer decoder \(\mathrm{Dec}\) refines these tokens through self-attention and cross-attention to \(\mathbf{H}_t^E\):
\[
\mathbf{H}_t^D \;=\; \mathrm{Dec}\bigl(\mathbf{A},\; \mathbf{H}_t^E\bigr).
\]
During cross attention, each action token attends to both vision and force representations. If we split \(\mathbf{H}_t^E\) into its vision (V) and force (F) parts, the cross-attention weights for each layer \(l\) can be decomposed as follows. For simplicity of notation, assume these weights are already averaged over multiple heads:

For the vision part:
\[
\alpha_{V}^{(l)} \;=\; \mathrm{softmax} \ \!\Bigl((\mathbf{A}^{(l)} \mathbf{W}^{Q(l)})\, (\mathbf{H}_{t,V}^{E(l)} \mathbf{W}^{K(l)})^\top / \sqrt{d}\Bigr),
\]

For the force part:
\[
\alpha_{F}^{(l)} \;=\; \mathrm{softmax} \ \!\Bigl((\mathbf{A}^{(l)} \mathbf{W}^{Q(l)})\, (\mathbf{H}_{t,F}^{E(l)} \mathbf{W}^{K(l)})^\top / \sqrt{d}\Bigr).
\]

These \(\alpha_{V}^{(l)}\) and \(\alpha_{F}^{(l)}\) measure how strongly each action token attends to vision vs.\ force tokens at layer \(l\), and will be the main source of analysis in Sec.~\ref{sec:analysis}.

Finally, we project the decoder output \(\mathbf{H}_t^D\) to action space, which represents joint position targets for the follower arm:
\[
\hat{q}_{t:t+k} 
= \mathrm{MLP}\bigl(\mathbf{H}_t^D\bigr) \;\in\; \mathbb{R}^{l\times d_a}.
\]
where \(d_a\) is the dimension of the action space. Substituting $\hat{q}_{t:t+k}$ into Eq.~\ref{eq:bcloss} gives the full BC objective. Please see Appendix~\ref{sec:architecture} for the detailed policy architecture and training hyperparameters.

\subsection{Force-Attending Curriculum} 

Through experiments, as shown in Sec.~\ref{sec:analysis} and Fig.~\ref{fig:test_analysis}, we found that naively concatenating force data to the policy observation during training often results in policies that neglect force input, failing to leverage force input to the fullest extent. To address this, we employ a \emph{curriculum} that gradually unveils detailed visual information, encouraging the model to learn to utilize force first. Specifically, we define two operators:
\(\beta_P(I,\, \sigma_n)\) for the pixel space, and \(\beta_L(z,\, \sigma_n)\) for the latent space, where \(\sigma_n\) is a scale parameter (e.g.\ the standard deviation of a Gaussian kernel or the kernel size of a max pooling operator) that is updated over the course of training for $N$ total gradient steps. During training, we apply the pixel-space operator $\beta_P$ to image $I_t$ or $\beta_L$ to visual latent tokens $z_t^V$.

Intuitively, the operators make visual inputs or latent tokens close in the metric space, thus encouraging more contribution from the force modality, particularly at the start of the training. Consider the limit $\sigma \to \infty$, each visual input converges to approximately the same tensor. Hence, the model can only learn a single global output for all visual inputs. Thus, at the early stage of the curriculum, the gradient updates focus more on using the force information and updating the force encoder to maximally differentiate between inputs.

\begin{algorithm}[t]
\captionof{algorithm}{Force-Attending Curriculum Training (FACTR)}
\label{alg:curriculum}
\begin{algorithmic}[1]
\small
\STATE \textbf{Given:} Expert dataset \(\mathcal{D}\); action chunking size $k$; total training steps \(N\); Pixel-space operator \(\beta_P(I,\, \sigma)\); latent-space operator \(\beta_L(z,\, \sigma)\); Scheduler defining \(\sigma_n\) for \(n=1\ldots N\)

\STATE Initialize pre-trained ViT \(f_\phi\), force MLP encoder \(g_\psi\), and action-chunking transformer \(\pi_\theta\)
\FOR{iteration \(n = 1 \dots N\)}
    \STATE Sample \(\bigl(I_t, \tau_t,\, q_t\bigr)\) from \(\mathcal{D}\)
    \STATE \(\sigma_n \leftarrow \text{Scheduler}(n)\) \quad // get current scale
    \IF {pixel-space curriculum}
    \STATE \(I_t \;\leftarrow\; \beta_P\bigl(I_t,\, \sigma_n\bigr)\) \quad 
    \ENDIF
    \STATE \(\mathbf{z}_t^V \;\leftarrow\; f_\phi(I_t)\) \quad // vision tokens
    \IF {latent-space curriculum}
    \STATE \(\mathbf{z}_t^V \;\leftarrow\; \beta_L\bigl(\mathbf{z}_t^V,\, \sigma_n\bigr)\)
    \ENDIF
    \STATE \(\mathbf{z}_t^F \;\leftarrow\; g_\psi(\tau_t)\)   \quad // force token
    \STATE \(\hat{q}_{t:t+k} \;\leftarrow\; \pi_\theta\bigl(\mathbf{z}_t^V,\; \mathbf{z}_t^F\bigr)\)
    \STATE \(\displaystyle \mathcal{L} \;=\; \mathrm{MSE}\!\bigl(\hat{q}_{t:t+k},\, q_{t:t+k}\bigr)\)
    \STATE Update \(\phi,\;\psi,\;\theta\) using ADAM
\ENDFOR
\end{algorithmic}
\end{algorithm}

\enlargethispage{1\baselineskip} 
\subsection{Curriculum Operators} 
\label{sec:operator}
We consider two types of operators: Gaussian blur and downsampling.

For the Gaussian blur, we define the 2D kernel \(G_{\sigma}\) as:
\[
G_{\sigma}(x, y) = \frac{1}{2\pi\sigma^2} \exp\!\left(-\frac{x^2 + y^2}{2\sigma^2}\right)
\]
The operator \(\beta_P(I, \sigma)\) applies this kernel using the 2D convolution operator \(\ast\):
\[
\beta_P(I, \sigma) = I \ast G_{\sigma}
\]

For the 1D Gaussian blur, the kernel \(g_{\sigma}\) is defined as:
\[
g_{\sigma}(x) = \frac{1}{\sqrt{2\pi}\sigma} \exp\!\left(-\frac{x^2}{2\sigma^2}\right)
\]
The operator \(\beta_L(z^V, \sigma)\) similarily applies this kernel using 1D convolution:
\[
\beta_L(z^V, \sigma) = z^V \ast g_{\sigma}
\]

For downsampling, we use MaxPool followed by nearest interpolation. In 2D, the pixel-space operator \(\beta_{P}(I)\) is:
\[
\beta_{P}(I) = \text{NearestInterp}(\text{MaxPool2D}(I))
\]

In 1D, the latent-space operator \(\beta_{I}(z^V)\) is the same except that a MaxPool1D is used.

By gradually reducing \(\sigma_n\), the curriculum ensures that the model focuses first on force tokens, and then incorporates visual information in the later stage of the training. This produces a policy \(\pi_\theta\) that more robustly fuses force and vision for control, alleviating the issue of overfitting to the vision modality.

\vspace{.07in}\noindent\textit{Theoretical Analysis}:$\quad$
We analyze the effects of Gaussian blur as an example of a curriculum operator through the framework of Neural Tangent Kernels (NTK)~\cite{jacot2018neural}. Although we consider a simple two-layer model here, the intuition applies to more complex architectures like vision transformers (ViTs), which have a more sophisticated NTK~\cite{boix2023can}. The formal theoretical analysis is presented in Appendix~\ref{sec:ntk}.

\subsection{Curriculum Schedulers}
\label{sec:scheduler}
Over the course of training (indexed by \(n=1,\dots,N\)), we adjust \(\sigma_n\) via a scheduler to control the information released from the visual branch. Given a initial scale \(\sigma_0\), we consider the following schedulers:

\begin{table}[h]
    \centering
    \renewcommand{\arraystretch}{1.5}
    \begin{tabular}{l l}
        \hline
        \textbf{Decay Type} & \textbf{Scheduler Equation} \\
        \hline
        \textit{Constant} & $\sigma_n = \sigma_0$ \\
        \textit{Linear} & $\sigma_n = \sigma_0 \left( 1 - \frac{n}{N} \right)$ \\
        \textit{Cosine} & $\sigma_n = \frac{\sigma_0}{2} \Bigl(1 + \cos\Bigl(\tfrac{n\pi}{N}\Bigr)\Bigr)$ \\
        \textit{Exponential} & $\sigma_n = \sigma_0 \cdot \alpha^n, \quad \alpha > 1$ \\
        \textit{Step} & $\sigma_n = \sigma_0 \Bigl(1 - \frac{1}{d_{\text{steps}}} \lfloor \frac{n}{N / d_{\text{steps}} } \rfloor\Bigr), \quad d_{\text{steps}} > 1$ \\
        \hline
    \end{tabular}
    \caption*{}
    \label{tab:decay_functions}
    \vspace{-8mm}
\end{table}

Furthermore, we warm-up the curriculum by fixing the scale to $\sigma_0$ for certain gradient steps, and adjust the decay formula to account for this duration. The rationale behind this step is to warm-up the randomly initialized force encoder with relatively low visual information. 

\begin{figure}
    \centering
    \includegraphics[width=\linewidth]{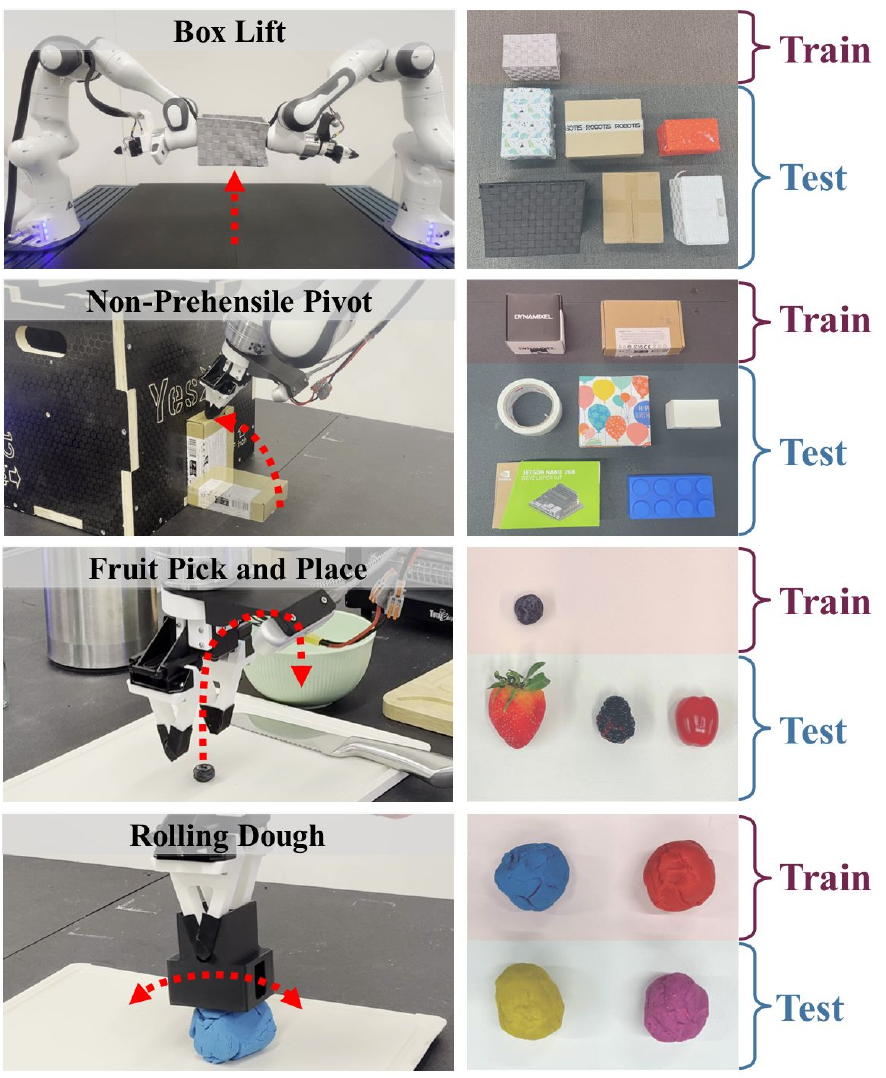}
    \caption{\small \textbf{Tasks.} We evaluate our leader-follower teleoperation system and autonomous policies trained with \our on four contact-rich tasks. These tasks are challenging as they require the robot to perceive and respond to the force feedback as it manipulates objects with unseen visual appearances and geometries. }
    \label{fig:task}
    \vspace{-5mm}
\end{figure}


\section{Evaluation}
\enlargethispage{1\baselineskip} 

\subsection{Experimental Setup}
We setup four contact-rich tasks, which are illustrated in Fig.~\ref{fig:task} along with the training and testing objects of various shapes and visual appearances. For all tasks, we use Franka Panda arm(s) with OpenManipulator-X gripper(s). Each task uses either a front ZED2 camera or wrist cameras mounted near the grippers with RGB observations. We describe the tasks details and the success criteria below.

\begin{itemize}
    \item \textit{Box Lift}: A bi-manual task where two arms lift a box and balance in the air for at least two seconds, using the front camera and external joint torque from the arms.
    \item \textit{Non-Prehensile Pivot}: The robot flips an item by pivoting it against the corner of a fixture until the item is rotated by $90^{\circ}$ and can stand stably, using the front camera and external joint torque from the arm.
    \item \textit{Fruit Pick and Place}: The robot grasps a soft and delicate fruit and places it in a bowl, using the wrist camera and the external joint torque of the gripper.
    \item \textit{Rolling Dough}: The robot continuously rolls the dough to shape it into a cylinder for at least 8 seconds, using the front camera and the external joint torque of the arm.
\end{itemize}

\subsection{Teleoperation Evaluation}
\begin{figure*}
    \centering
    \includegraphics[width=\linewidth]{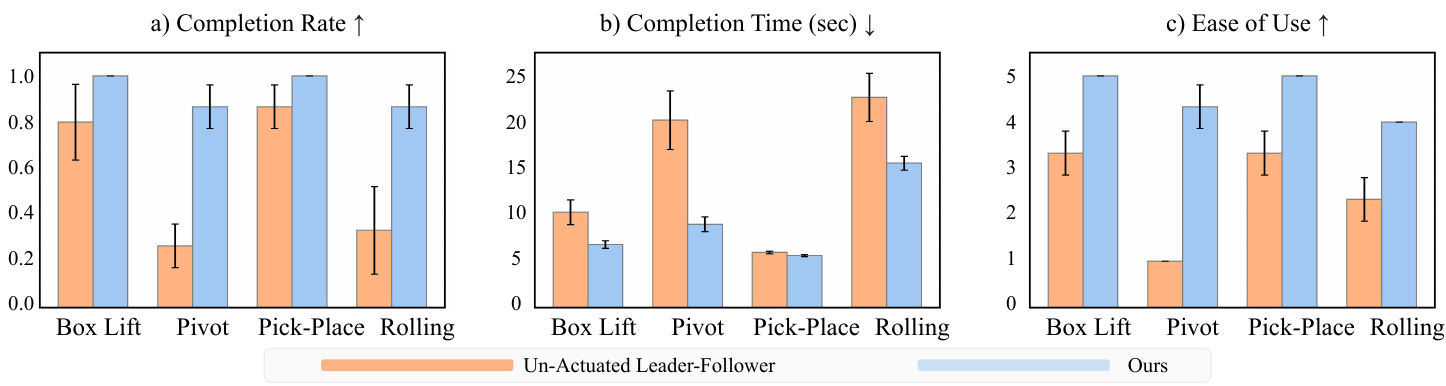}
    \caption{\small\textbf{User study.} \our teleoperation system allows users to complete tasks with significantly higher success rate, using less time, and they subjectively found our system to be easier to use.}
    \label{fig:user_study}
    \vspace{-4mm}
\end{figure*}

We compare our leader-follower teleoperation system, which includes our leader arm with force feedback, gravity compensation, and redundancy resolution, to an un-actuated leader-follower baseline system with mechanical joint regulation, similar to \cite{wu2024gello}. We summarize our results in Fig.~\ref{fig:user_study}.

Our experiments show that our system allows users to complete tasks with 64.7\% higher task completion rate, 37.4\% reduced completion time, and 83.3\% improvement in the subjective ease of use metrics.

We observe that for tasks that require continuous contact between the arm and an object, such as non-prehensile pivoting and bimanual box lifting, the un-actuated teleoperation system often causes the follower arm to lose contact with the object. This occurs because of the absence of force feedback, which prevents the user from perceiving the environment's geometric constraints through the leader arms. As a result, maintaining continuous contact with the object becomes challenging.

For the un-actuated system, the follower arm frequently exceeds its joint velocity limits when moving under continuous contact. This occurs because the operator can easily maneuver the leader arms in ways that cause significant deviations between the leader and follower joint positions, especially when the follower arm is in contact with the environment. When contact is lost, the resulting large joint-space error causes the PID controller to generate large torques, causing abrupt movements that exceed the velocity limits. On the other hand, our system's force feedback renders geometric constraints of the environment for the operator through the leader arms, preventing the operator from moving the leader arms too far away from the follower arms during environment contacts.

\subsection{Policy Evaluation}

\textbf{Questions.} In our real-world evaluation, we seek to address the following research questions regarding \our: 
\begin{itemize}
    \item How does \our perform compared to baseline approaches that do not use force feedback and ones that use force feedback without \our?
    \item How do different curriculum parameters affect policy performance?
\end{itemize}

\begin{figure}[t]
    \centering
    \includegraphics[width=\linewidth]{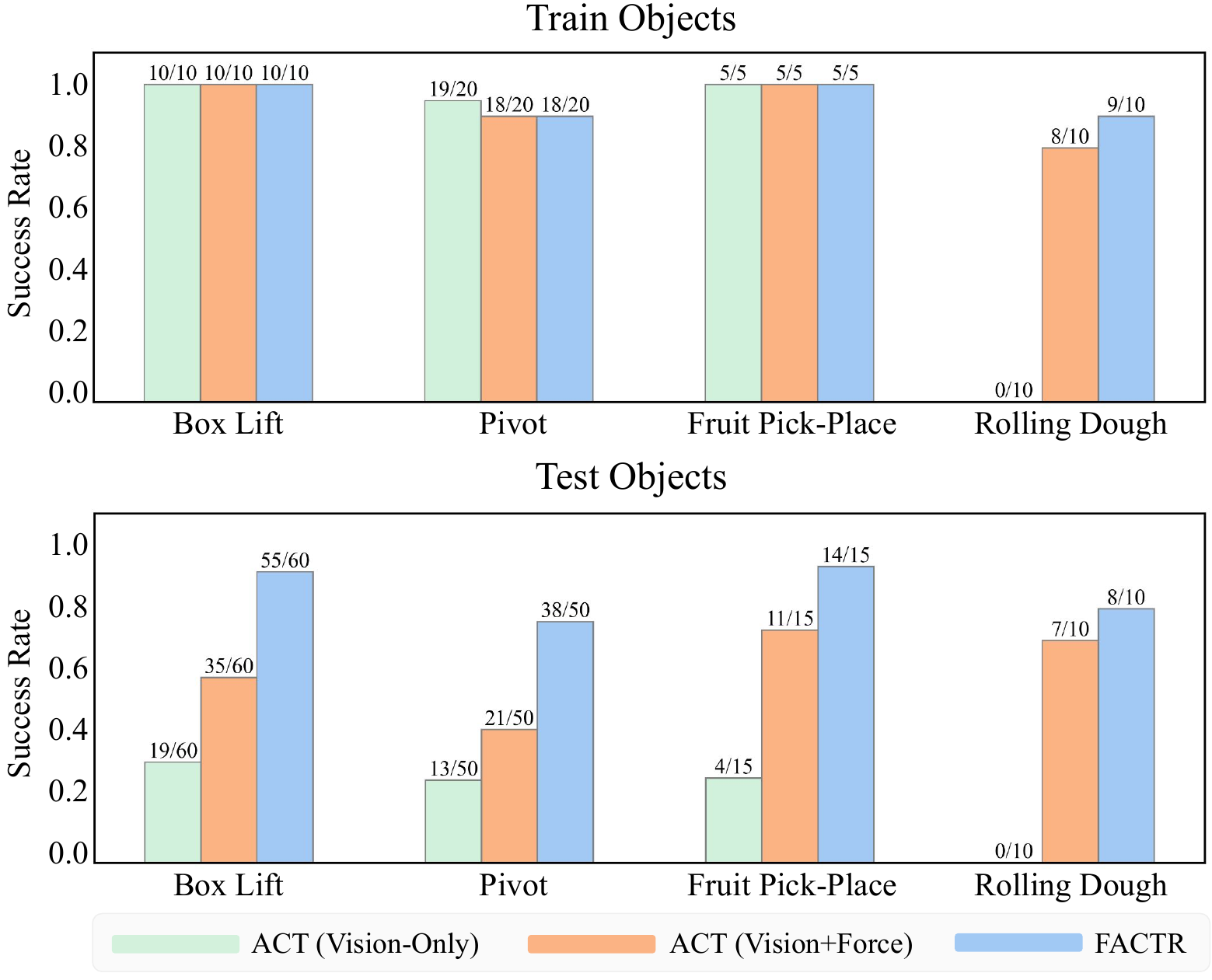}
    \caption{\small {FACTR leads to better object generalization.}}
    \label{fig:results}
    \vspace{-5mm}
\end{figure}

\textbf{Training and Evaluation Protocol.} We collected 50 demonstrations with our teleoperation system. We trained each method with the same hyperparameters, where details can be found in the Appendix~\ref{sec:architecture}. We compare the following methods:
\begin{itemize}
    \item \emph{ACT (Vision-Only)~\cite{aloha}}: Action Chunking Transformer which only takes in visual observation.
    \item \emph{ACT (Vision+Force)~\cite{kobayashi2024alpha, li2024haptic}}: Action Chunking Transformer which takes in both visual and force observation, but trained without a curriculum.
    \item \emph{FACTR (Ours)}: Action Chunking Transformer trained with Force-Attending Curriculum. For each task, we train a latent space curriculum with the Guassian Blur operator and linear scheduler. We discuss more detailed ablations on the curriculum in Sec.~\ref{sec:ablations}.
\end{itemize}

For each object in each task, we evaluated 5-10 trials. We present the average success rate for training and testing objects, respectively. Detailed evaluation results for each object can be found in the Appendix~\ref{sec:detailed_results}.

\begin{figure}[t]
    \centering
    \includegraphics[width=\linewidth]{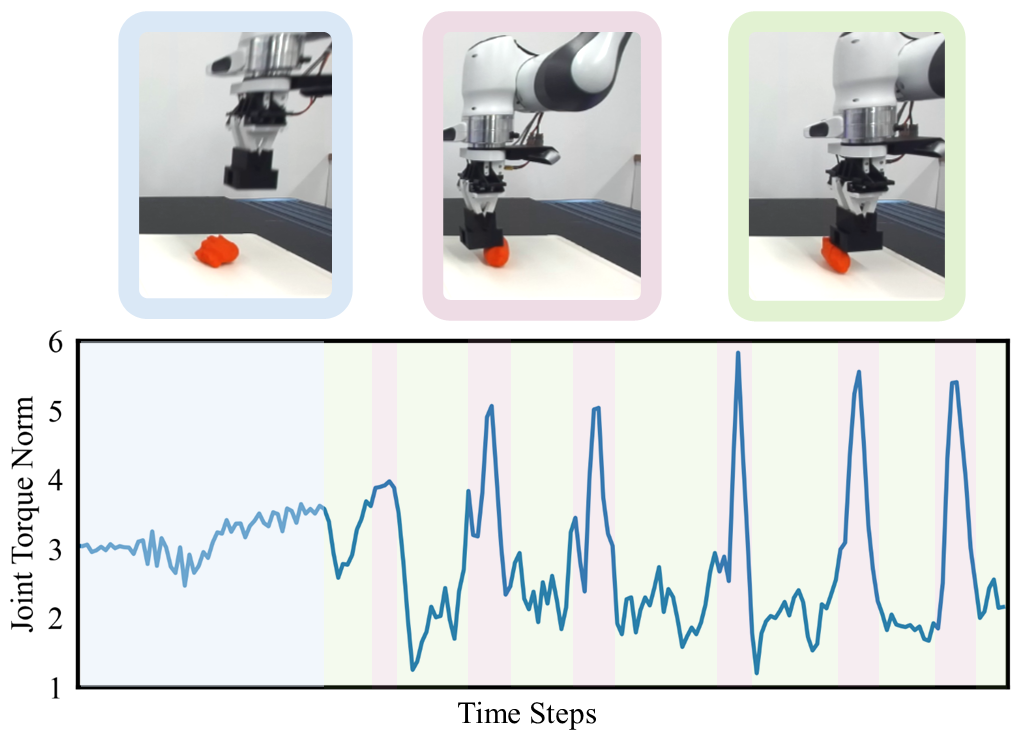}
    \caption{\small We visualize the external joint torque norm of the Franka arm for a collected trajectory. Blue highlights indicate pre-contact phases, while purple and green mark torque peaks and troughs at the dough’s left and right ends, respectively. The oscillatory torque pattern helps the policy distinguish observations despite similar visual inputs.
    }
    \label{fig:roll}
    \vspace{-5mm}
\end{figure}

\textbf{FACTR leads to better generalization.} We present our main quantitative results in Fig.~\ref{fig:results}. 
All the policies perform similarly on the train objects for most tasks, except for the rolling dough task, where the vision-only policy smashes the dough without any rolling actions and fails completely. Note that the visual observations are hard to distinguish during the oscillatory rolling motions, while the force signals form a corresponding oscillatory pattern, as shown in Fig.~\ref{fig:roll}; this distinctive torque pattern helps policies with force input to complete the task.

For the test objects, the vision-only policy achieves a success rate of 21.3\% on average, which is significantly worse than policies incorporating force. Without a curriculum, policies naively incorporating force achieve a success rate of 61.2\%, while \our achieves a success rate of 87.5\%, which shows that \our leads to significantly better generalization to novel objects. We hypothesize that the force information provides important signals for mode switching at moments such as when the robots get into contact with the box in the lifting task and when the object is grasped in the fruit pickup task.

\textbf{Policies with FACTR learns to identify mode switching.} 
\label{sec:analysis}
To better understand the policies trained with \our. We visualize the attention behavior during policy training and inference. Specifically, we visualize the cross attention of the action tokens to the memory tokens denoted as $\alpha_{V}^{(1)}$ and $\alpha_{F}^{(1)}$ for the first layer of the decoder, where $\alpha_{V}^{(1)}$ and $\alpha_{F}^{(1)}$ are defined in Sec.~\ref{sec:method}.

\begin{figure*}
    \centering
    \vspace{-1mm}
    \includegraphics[width=.96\linewidth]{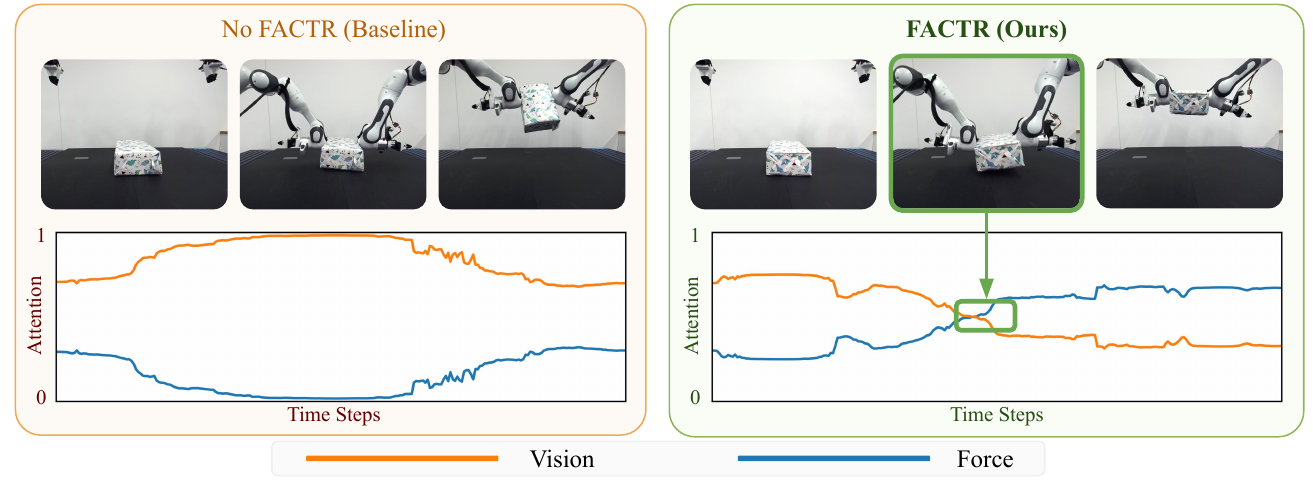}
    \vspace{-1mm}
    \caption{\small
        \textbf{Policies trained with \our learns to identify mode switching.} We visualize the average cross attention of the action tokens to the \textcolor{blue}{force} or \textcolor{orange}{vision} tokens of the first decoder layer during policy rollout. [Left] Without the curriculum, the policy does not pay enough attention to force, and either fails to lift or balance the novel boxes. [Right] \our learns to attend to force more to complete the task. For example, in the box lifting task, attention to force outweighs that of vision as the arms contact the box, signaling a mode switch.  
    }
    \label{fig:test_analysis}
    \vspace{-6mm}
\end{figure*}

During policy rollout, we visualize the average cross attention of the action tokens to the \textcolor{blue}{force} or \textcolor{orange}{vision} tokens of the first decoder layer as shown in Fig.~\ref{fig:test_analysis}. \our learns to attend to force more during task execution. For example, in the box lifting task, attention to force outweighs that of vision as the arms contact the box, signaling a mode switch. While without the curriculum, the policy does not pay enough attention to force, and either fails to lift or balance the novel boxes.

\textbf{FACTR leads to better recovery behavior.}
Another notable observation is that FACTR also facilitates recovery behavior. Specifically, we evaluate the box-lifting task with five trials per object. A trial begins when the policy successfully lifts the box for the first time; we then knock the box down and assess the second attempt. As shown in Table~\ref{tab:recovery}, all policies maintain nearly 100\% recovery success on training objects. However, for test objects, the vision-only policy's success rate drops significantly from 31.7\% on the first attempt to 13.3\% on the second. In contrast, force-attending policies maintain similar success rates across both attempts.

We observe that vision-only policies often remain static after the box is knocked down, failing to retry. We hypothesize that this occurs because the vision-only policy overfits to training scenarios, making it unresponsive to unseen objects outside its training distribution. In contrast, \our policies detect loss of contact through external joint torque readings, which revert to pre-lift values when the object is dropped. Since our \our policies effectively attend to force input, they successfully recover to a pre-lift state and attempt the task again.

\begin{table}[h]
    \centering
    \begin{tabular}{l c c}
        \toprule
        & Train Objects & Test Objects \\
        \midrule
        ACT (Vision-Only) & 4/5 & 4/30 \\
        ACT (Vision+Force) & 5/5 & 16/30 \\
        \midrule
        \textbf{FACTR} & \textbf{3/3} & \textbf{27/30} \\
        \bottomrule
    \end{tabular}
    \caption{\small Evaluation of recovery behaviors for box lifting.}
    \label{tab:recovery}
    \vspace{-4mm}
\end{table}

\subsection{Ablations on Curriculum}
\label{sec:ablations}

To further validate the significance of a curriculum, we trained models with fixed $\sigma_n$ across training. Moreover, to ablate on pixel space and latent space curriculum, and different scheduler and operator choices, we train policies on different combination of these parameters. We choose the task of \textit{pivoting}, one of the hardest tasks from our task suite, for the ablations. We evaluate only on the five \emph{test objects} for five trials each, since they are more indicative of policy performance than train objects. The results are presented in TABLE~\ref{tab:ablation}.

\begin{table}[h]
\vspace{-2mm}
\centering
\begin{tabular}{lcccc}
\toprule
& \multicolumn{2}{c}{Pixel Space} & \multicolumn{2}{c}{Latent Space} \\
\cmidrule(lr){2-3} \cmidrule(lr){4-5}
& Blur & Downsample & Blur & Downsample \\
\midrule
Constant & 16/25 & 15/25 & 17/25 & 16/25 \\
\midrule
Linear & 19/25 & 18/25 & 19/25 & 18/25 \\
Cosine & 20/25 & 19/25 & 17/25 & 19/25 \\
Exp & 19/25 & 21/25 & 20/25 & 19/25 \\
Step & 19/25 & 18/25 & 20/25 & 19/25 \\
\bottomrule
\end{tabular}
\caption{\small\textbf{Curriculum ablation.}}
\label{tab:ablation}
\vspace{-3mm}
\end{table}

\textbf{Fixed-Scale Operator vs. Curriculum.} 
We found that performance with a curriculum of decaying smoothing performs better than a fixed curriculum across all tasks. We hypothesize that to enable better performance, the final policy needs to take in the fully unblurred vision information. Through the curriculum, a policy gets to gradually adapt to unblurred images. On the other hand, with fixed smoothing, even though policy may not overfit to visual information, it cannot extract the necessary details from unblurred vision to complete the tasks.

\textbf{Comparisons with other scheduler parameters.} 
We further compare policies trained with either pixel space or latent space, two operators (Gaussian blur and downsample) as defined in Sec.~\ref{sec:operator}, and four schedulers (linear, cosine, exponential, and step) as defined in Sec.~\ref{sec:scheduler}. However, we do not find a uniform advantage or disadvantage for any set of parameters. The results suggest that FACTR is relatively robust to different sets of curriculum parameters.

\section{Conclusion and Limitations} 
\label{sec:conclusion}
We introduced FACTR, a curriculum approach to train force-based policies to improve performance and object generalization in contact-rich tasks. FACTR leverages a blurring operator with decreasing scales on the visual information throughout training. This encourages the policy to leverage force input at the beginning stages of training, preventing the problem where the policy overfits to visual input and thus neglects force input. This approach was demonstrated through a series of experiments on the following tasks: box lifting, non-prehensile pivoting, fruit pick-and-place, and rolling dough, where FACTR exhibits significant improvements in task completion rates and generalization to unseen object appearances and geometries. Additionally, our teleoperation system, which includes an actuated leader arm for force feedback and gravity compensation, was shown to provide a more intuitive user experience, as evidenced by higher task completion rates and user satisfaction in our studies.

While FACTR demonstrates significant improvements in force-based policy learning for contact-rich tasks, it has limitations. First, the precision of the external joint torque sensors in our follower arm is limited. This limitation can particularly affect tasks that involve subtle force adjustments during fine-grained manipulation since the torque readings can be too noisy to be used effectively. Future work could explore integrating high-resolution tactile sensors or haptic gloves to enhance feedback precision and improve overall system performance. Second, our approach assumes the availability of external joint torque sensors in the follower arms. Future work can explore adapting our system for an arm mounted with an end-effector force-torque sensor. Third, the effectiveness of our curriculum learning approach can be influenced by several hyperparameters, such as the choice of the blurring operator and scheduling strategies. These parameters can be highly task-dependent, requiring extensive tuning for different applications. Developing adaptive or self-tuning curriculum strategies could help mitigate this issue by dynamically adjusting hyperparameters based on task-specific requirements. Addressing these limitations could further enhance FACTR’s applicability and robustness across a broader range of contact-rich manipulation tasks.

\section*{Acknowledgments}
We thank Arthur Allshire, Andrew Wang, Mohan Kumar Srirama, Ritvik Singh for discussions about the paper. We also thank Tiffany Tse, Ray Liu, Sri Anumakonda, Sheqi Zhang with teleoperation. This work is supported in part by ONR MURI N00014-22-1-2773, ONR MURI N00014-24-1-2748 and AFOSR FA9550-23-1-0747.

\bibliographystyle{plain}
\bibliography{references}

\clearpage

\section*{\centering Appendix}

\section{Analysis of FACTR from Neural Tangent Kernel (NTK) Perspective}
\label{sec:ntk}

In this section, we aim to give a sketch of a more theoretical argument for why operators like Gaussian blur or downsampling in FACTR could help the policy attending to force input. We will first briefly introduce Neural Tangent Kernel (NTK), and then use it a theoretical framework to analyze the effects of Gaussian blur as an example curriculum operator.

\subsection{Preliminaries on Neural Tangent Kernel (NTK)}

The Neural Tangent Kernel (NTK) is a theoretical framework used to analyze the behavior of neural networks, particularly in the limit of infinite width~\cite{jacot2018neural, arora2019exact}. For a neural network \(f_\theta(x)\) with parameters \(\theta\), the NTK is defined as:
\[
k(x_i, x_j) = \langle \nabla_\theta f_\theta(x_i), \nabla_\theta f_\theta(x_j) \rangle,
\]

where \(\nabla_\theta f_\theta(x)\) is the gradient of the network's output with respect to its parameters \(\theta\), and \(\langle \cdot, \cdot \rangle\) denotes the inner product. In the infinite width limit, the NTK becomes deterministic and remains constant during training. Assuming the parameters \(\theta\) are initialized from a Gaussian distribution, the NTK \(k(x_i, x_j)\) converges to a deterministic kernel \(k_\infty(x_i, x_j)\) given by:
\[
k_\infty(x_i, x_j) = \mathbb{E}_{\theta \sim \mathcal{N}(0, I)} \left[ \langle \nabla_\theta f_\theta(x_i), \nabla_\theta f_\theta(x_j) \rangle \right],
\]

where the expectation is taken over the Gaussian initialization of \(\theta\).

The NTK \(k(x_i, x_j)\) is a similarity function: if \(k(x_i, x_j)\) is large, then the predictions \(f_\theta(x_i)\) and \(f_\theta(x_j)\) will tend to be close. If we have \(n\) training points \((x_i, y_i)\), \(k\) defines a positive semi-definite (PSD) kernel matrix \(K \in \mathbb{R}^{n \times n}_+\) where each entry \(K_{ij} = k(x_i, x_j)\).

Fascinatingly, when we train this infinite-width neural network with gradient flow on the squared error, we precisely know the model output at any point in training. At time \(t\), the training residual is given by:

\[
f_{\theta_t}(x) - y = e^{-\eta K t} (f_{\theta_0}(x) - y),
\]

where \(\eta\) is the learning rate, \(K\) is the kernel matrix, and \(f_{\theta_0}(x)\) is the initial model output. This equation shows that the residual error decays exponentially with a rate determined by the kernel matrix \(K\).

\subsection{Analyzing the Effects of Gaussian Blur with NTK}

We analyze the effects of Gaussian blur as an example of a curriculum operator. While we consider a simple two-layer model here, the intuition applies to more complex architectures like vision transformers (ViTs), which have a more sophisticated NTK~\cite{boix2023can}.

Consider a model where the input \(x\) is first convolved with a Gaussian kernel \(K_\sigma\) before being passed through the network, where \(\sigma\) is the standard deviation of the Gaussian kernel. The output of the model is:

\[
f(x) = W^T (K_\sigma * x),
\]

where \(K_\sigma * x\) is the convolution of \(x\) with a Gaussian kernel \(K_\sigma\). Assuming that the model has infinite width and the parameters in \(W\) are initialized from a Gaussian distribution, the NTK for this model is:

\begin{align*}    
    k_\sigma(x, x') &= \mathbb{E}_{W \sim \mathcal{N}(0, I)} \left[ \langle \nabla_W f(x), \nabla_W f(x') \rangle \right] \\
    &= \mathbb{E}_{W \sim \mathcal{N}(0, I)} \langle K_\sigma * x, K_\sigma * x' \rangle.
\end{align*}

This is the dot product between the convolved inputs \(K_\sigma * x\) and \(K_\sigma * x'\). As \(\sigma\) increases, the Gaussian convolution \(K_\sigma * x\) acts as a low-pass filter, attenuating high-frequency components in the input \(x\). This causes the convolved inputs \(K_\sigma * x_i\) and \(K_\sigma * x_j\) to become more similar for any pair of inputs \(x_i\) and \(x_j\). In the extreme case where \(\sigma \to \infty\), \(K_\sigma * x_i \approx K_\sigma * x_j\) for all \(x_i, x_j\), and the NTK \(k_\sigma(x_i, x_j)\) approaches a constant value. This implies that the kernel matrix \(K\) becomes approximately an all-ones matrix, scaled by the magnitude of the convolved inputs.

In our curriculum, we decrease \(\sigma\), where the model is first exposed to smoother visual data before gradually transitioning to the original unsmoothed data. At the beginning of the curriculum, the model focuses on low-frequency patterns and robust features from visual inputs. As \(\sigma\) decreases to small \(\sigma\), the NTK retains discriminative power, allowing the model to learn fine-grained features from visual inputs. . This gradual increase in complexity can help the model learn more effectively by avoiding overfitting to the high-frequency variations in visual inputs in the early stages of training.

More rigorously, consider the limit $\sigma \to \infty$, each blurred input $K_{\sigma} * x_i$ converges to the same vector $\phi$. Hence,
\[
k_\sigma(x_i, x_j) \;=\; \langle \phi, \phi \rangle \;=\; \|\phi\|^2,
\]
and the NTK matrix $K$ becomes
\[
K \;=\; \|\phi\|^2
\begin{pmatrix}
1 & 1 & \cdots & 1\\
1 & 1 & \cdots & 1\\
\vdots & \vdots & \ddots & \vdots\\
1 & 1 & \cdots & 1
\end{pmatrix}.
\]
This matrix is rank-1, with largest eigenvalue $\lambda = n \|\phi\|^2$ (for $n$ training points) and corresponding eigenvector $v = (1,1,\dots,1)$.

Recall that in the infinite-width NTK regime, the training residual $r$ satisfies
\[
r(t) \;=\; e^{-\eta K t}\,r(0),
\]
where $r(0)=f_{\theta_0}(x) - y$ is the initial residual. Decompose $r(0)$ into components parallel and perpendicular to $v$:
\[
r(0) \;=\; r_{\parallel} + r_{\perp}, \quad
\text{with } r_{\parallel} \propto v \text{ and } r_{\perp} \cdot v = 0.
\]
Since $K$ is rank-1,
\[
e^{-\eta K t}\,r(0)
\;=\; r_{\perp} \;+\; e^{-\eta \lambda t}\,r_{\parallel}.
\]
The parallel component decays exponentially at rate $\lambda$, while the perpendicular component is unchanged. This effectively learns a single global scalar for all inputs, reducing mean-squared error by matching the average label but losing discriminative power. Thus, at the early state of the curriculum, the gradient updates will focus on using the force information and updating the force encoder to maximally differentiate between inputs.

\section{Cost Analysis of Our Teleoperation System with Force Feedback}
\label{sec:cost}

Please see Table~\ref{tab:cost} for a detailed Bill of Materials and breakdown of the cost to create one leader arm and leader gripper as part of our teleoperation with force feedback system. This is accurate pricing as of the paper release.

\begin{table}[h]
\centering
\begin{tabular}{lcc}
    \toprule
        Object & Quantity & Total \\
     \midrule
        Dynamixel XM430-W210-T & 2 & \$539.80 \\
        Dynamixel XC330-T288-T & 6 & \$539.40 \\
        U2D2 Control PCB & 1 & \$32.10 \\
        12V 20A Power Supply & 1 & \$24.99 \\
        FPX330-S102 Servo Bracket & 1 & \$8.70 \\
        Polymaker PLA PRO Filament & 1 & \$24.99 \\
        U2D2 Power Hub Board & 1 & \$19.99 \\
        14AWG Cable & 1 & \$23.99 \\
        3/4" Bearing & 1 & \$6.99 \\
        Screws & - & \$15.99 \\
     \midrule
        Total & & \$1229.95 \\
    \bottomrule
\end{tabular}
\vspace{0.1in}
\caption{\small We present the bill of materials of one leader arm teleoperation device with force feedback. The total cost is around \$1229.95.}
\label{tab:cost}
\vspace{-0.1in}
\end{table}

\section{Additional Control Laws for Our Teleoperation System}
\label{sec:add_control_law}

Here we define the additional control laws for our teleoperation system with force feedback: friction compensation and joint limit avoidance.

\subsection{Friction Compensation}
Joint friction introduces resistive forces that impair responsiveness, making precise motion control challenging and reducing the system's intuitive feel for the operator \cite{friction_bad}. Friction also increases the physical effort required to back-drive the motor, leading to operator fatigue during prolonged use. To this end, we use a dynamic friction model to explicitly compensate for static, Coulomb, and viscous friction \cite{friction_models}. 

We mitigate the effects of static friction by introducing a small, high-frequency oscillatory signal to the control input, which generates micro-vibrations that prevent the motors from settling into static friction states, thereby enabling smoother transitions from rest to motion \cite{friction_models}. The static friction compensation torque $\tau_{ss}$ is as follows:
\begin{equation}
    \tau_{ss}^{(i)} =\begin{cases} 
        \mu_s^{(i)} \cos(\frac{\pi t}{f}) & \text{if } \dot{q}^{(i)} < \dot{q}^{(i)}_{s}, \\
        0 & \text{otherwise}
    \end{cases}
\end{equation}
where $\mu_s$ is a calibrated static friction coefficient and $f$ is the control loop frequency, set to 500 Hz.

We also account for kinetic friction by compensating for Coulomb friction and viscous friction as follows \cite{friction_models}:
\begin{equation}
    \tau_{ks}^{(i)} = \mu_{c}^{(i)}\text{sgn}(\dot{q}^{(i)}) + \mu_v^{(i)} \dot{q}^{(i)}
\end{equation}
where $\mu_c$ is the Coulomb friction coefficient and $\mu_v$ is viscous friction coefficient.

The total friction compensation $\tau_{friction}$ is the sum of the static friction $\tau_{ss}$ and kinetic friction $\tau_{ks}$ terms.

\subsection{Joint Limit Avoidance}
We implement the following artificial potential-based control law to prevent the operator from making the leader arm go beyond the joint limits of the follower arm:
\begin{equation}
    U(q^{(i)}) =
    \begin{cases}
    \frac{1}{2} \eta \frac{1}{(q^{(i)} - q_{\text{min}}^{(i)})^2}, & q^{(i)} < q_{\text{min}}^{(i)} + \Delta q \\
    \frac{1}{2} \eta \frac{1}{(q_{\text{max}}^{(i)} - q^{(i)})^2}, & q^{(i)} > q_{\text{max}}^{(i)} - \Delta q \\
    0, & \text{otherwise}
    \end{cases}
\end{equation}
\begin{equation}
    \tau_{limit}^{(i)} = -\nabla_{q^{(i)}}U(q^{(i)})
\end{equation}
where $U(q^{(i)})$ is the repulsive potential function, $\Delta q$ is the safety margin, and $\eta$ is the scaling factor. 

\subsection{Bi-manual Follower Arms Control with Dynamic Collision Avoidance}
Most existing bi-manual teleoperation systems with a leader-follower setup command the follower arms by directly setting the joint position targets to the current joint positions of the leader arm. Instead, we employ a Riemannian Motion Policy (RMP) \cite{RMP} implemented in Isaac Lab \cite{mittal2023orbit}, where the RMP dynamically generates joint-space targets for the follower arms that best match the current joint positions of the leader arms while incorporating real-time collision avoidance. Our system prevents the follower arms from colliding with one another or with external obstacles, such as the table, regardless of the operator's actions.

\section{Behavior Cloning Policy Architecture and Training Hyperparameters}
\label{sec:architecture}

Our behavior cloning policy takes as input a RGB image and current hand joint angles (proprioception). We obtain tokens for the image observation via a ViT~\cite{vit} and a token for joint proprioception via a linear layer. The weights of ViT is initialized from the Soup 1M model from~\cite{dasari2023unbiased}. The tokens then pass through action chunking transformer, an encoder-decoder transformer, to output a sequence of actions~\cite{aloha}. The action space is the absolute joint angles of the two arms for box lift, the absolute angles of a single arm for non-prehensile pivot and rolling dough, and the absolute angles of a arm and the gripper for fruit pick and place.  A key decision that greatly improves policy generalization is to exclude current arm joints from the proprioception. Intuitively, this may force the model to extract object information from image observations, rather than overfitting to predict actions close to current arm states.

We list key hyperparameters for our behavior policy training Table~\ref{tab:hyperparam}. In general, we are able to obtain well-performing policies with 20000-50000 gradient steps and 2-6 hours of wall-clock time training on a RTX4090.

\begin{table}[t]
\centering
\begin{tabular}{@{}p{3.5cm}p{4cm}@{}}
\toprule
\textbf{Hyperparameter} & \textbf{Value} \\
\midrule
\multicolumn{2}{c}{\textbf{Behavior Policy Training}} \\
\midrule
Optimizer & AdamW \\
Base Learning Rate & 3e-4 \\
Weight Decay & 0.05 \\
Optimizer Momentum & $\beta_1, \beta_2 = 0.9, 0.95$ \\
Batch Size & 128 \\
Learning Rate Schedule & Cosine Decay \\
Total Steps & 20000-50000 \\
Warmup Steps & 500 \\
Augmentation & RandomResizeCrop \\
GPU & RTX4090 (24 gb) \\
Wall-Clock Time & 2-6 hours\\
\midrule
\multicolumn{2}{c}{\textbf{Visual Backbone ViT Architecture}} \\
\midrule
Patch Size & 16 \\
\# Layers & 12 \\
\# MHSA Heads & 12 \\
Hidden Dim & 768 \\
Class Token & Yes \\
Positional Encoding & sin cos \\
\midrule
\multicolumn{2}{c}{\textbf{Action Chunking Transformer Architecture}} \\
\midrule
\# Encoder Layers & 6 \\
\# Decoder Layers & 6 \\
\# MHSA Heads & 8 \\
Hidden Dim & 512 \\
Feed-Forward Dim & 2048 \\
Dropout & 0.1 \\
Positional Encoding & sin cos \\
Action Chunk & 100 \\
\bottomrule
\end{tabular}
\vspace{2mm}
\caption{\small Policy Architecture and Training Hyperparameters}\label{tab:hyperparam}
\vspace{-4mm}
\end{table}

\section{Additional Experiments}

\textbf{Adaptive Layer Norm as an Alternative to a Curriculum}
As an alternative to a curriculum, we experimented with Adaptive Normalization Layers~\cite{peebles2023scalable}
on the action tokens conditioned on the force input to improve the force conditioning (AdaNorm in Tab.~\ref{tab:add_baselines}). However, we found that it creates instability in training and leads to overfitting.

\textbf{Data Augmentation as alternative to curriculum.} We run 10 data augmentation experiments varying noise probabilities and noise levels applied to vision input. The best of these policies, shown as DataAug in Tab.~\ref{tab:add_baselines}. We find that policies trained with lower augmentation levels perform well on train objects but worse than FACTR on test objects, likely because low levels of augmentations fail to facilitate proper attention to force. Higher levels of augmentation policies do not perform well across train and test sets, likely because they fail to leverage vision altogether. Our curriculum approach eliminates such tradeoff between vision and force, allowing the policy to attend properly to both force and vision.

\begin{table}[t]
    \centering
    \footnotesize 
    \setlength{\tabcolsep}{3pt}
    \begin{tabular}{l@{\hskip 4pt} c@{\hskip 4pt} c@{\hskip 4pt} c@{\hskip 4pt} c@{\hskip 4pt} c@{\hskip 4pt} c}
        \toprule
        & \textbf{FACTR} & AdaNorm & NoiseAug \\
        \midrule
        Train (\%) & \textbf{90.0} & 25.0 & 85.0 \\
        Test (\%) & \textbf{77.7} & 6.1 & 65.0\\
        \bottomrule
    \end{tabular}
    \vspace{-1mm}
    \caption{\small Pivot task training and testing performance.}
    \label{tab:add_baselines}
\end{table}

\begin{figure}[t!]
    \centering
    \includegraphics[width=0.93\linewidth]{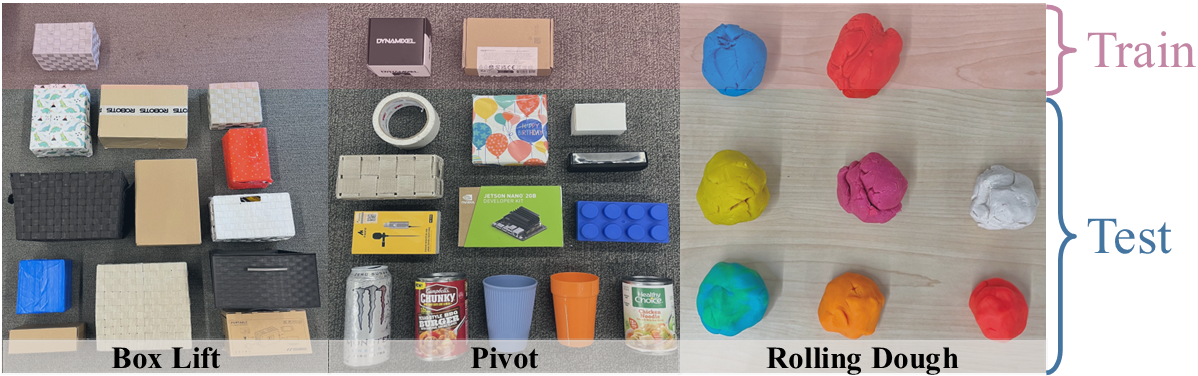}
    \vspace{-1mm}
    \caption{\small We more than doubled the testing set size across 3 tasks.}
    \label{fig:objects}
    \vspace{-5mm}
\end{figure}

\textbf{More test objects.}
We doubled the number of testing objects used in three of our tasks, visualized in Fig.~\ref{fig:objects} and updated the success rate comparison figure in Tab.~\ref{tab:generalization}. By conducting testing on more unseen objects and comparing with the best performing baseline (Bi-ACT), we observe that our method yields an improved generalization performance.

\begin{table}[t!]
    \centering
    \footnotesize
    \setlength{\tabcolsep}{6pt}
    \begin{tabular}{l c c c}
        \toprule
        & Box Lift & Pivot & Rolling Dough \\
        \midrule
        ACT (Vision-Only) & 35/120 & 30/130 & 0/60 \\
        Bi-ACT & 68/120 & 76/130 & 41/60 \\
        \midrule
        \textbf{FACTR (Ours)} & \textbf{105/120} & \textbf{101/130} & \textbf{46/60} \\
        \bottomrule
    \end{tabular}
    \vspace{-1mm}
    \caption{\small Policy evaluation on unseen objects across 3 tasks.}
    \label{tab:generalization}
\end{table}

\section{Detailed Quantitative Results} 
\label{sec:detailed_results}

We present the detailed evaluation results for each task in TABLE~\ref{tab:box_lift}, \ref{tab:pivot}, \ref{tab:fruit_pick_place}, and \ref{tab:rolling_dough}.

\begin{table*}[t]
\centering
\resizebox{0.8\linewidth}{!}{%
\begin{tabular}{lccccccccc}
\toprule
& \multicolumn{1}{c}{\makebox[0pt][c]{Train}} & \multicolumn{6}{c}{\makebox[0pt][c]{Test}} & Train Avg & Test Avg \\
\cmidrule(lr){2-2} \cmidrule(lr){3-8} 
& Box1 & Box2 & Box3 & Box4 & Box5 & Box6 & Box7 &  & \\
\midrule
ACT (Vision-Only) & 10/10 & 7/10 & 1/10 & 1/10 & 6/10 & 3/10 & 1/10 & 100.0\% & 31.7\% \\
ACT (Vision+Force) & 10/10 & 2/10 & 4/10 & 4/10 & 10/10 & 10/10 & 5/10 & 100.0\% & 58.3\% \\
\midrule
FACTR & 10/10 & 8/10 & 7/10 & 10/10 & 10/10 & 10/10 & 10/10 & 100.0\% & 91.7\% \\
\bottomrule
\end{tabular}}
\caption{\small Comparison of methods for Box Lift task.}
\label{tab:box_lift}
\end{table*}

\begin{table*}[t]
\centering
\resizebox{0.8\linewidth}{!}{%
\begin{tabular}{lccccccccc}
\toprule
& \multicolumn{2}{c}{\makebox[0pt][c]{Train}} & \multicolumn{5}{c}{\makebox[0pt][c]{Test}} & Train Avg & Test Avg \\
\cmidrule(lr){2-3} \cmidrule(lr){4-8} 
& Box1 & Box2 & Box3 & Box4 & Box5 & Box6 & Box7 &  & \\
\midrule
ACT (Vision-Only) & 10/10 & 9/10 & 3/10 & 0/10 & 7/10 & 2/10 & 1/10 & 95.0\% & 26.0\% \\
ACT (Vision+Force) & 9/10 & 9/10 & 1/10 & 2/10 & 9/10 & 4/10 & 5/10 & 90.0\% & 42.0\% \\
\midrule
FACTR & 9/10 & 9/10 & 6/10 & 5/10 & 10/10 & 7/10 & 10/10 & 90.0\% & 76.0\% \\
\bottomrule
\end{tabular}}
\caption{\small Comparison of methods for Non-Prehensile Pivot task.}
\label{tab:pivot}
\end{table*}

\begin{table*}[t]
\centering
\resizebox{0.6\linewidth}{!}{%
\begin{tabular}{lcccccc}
\toprule
& \multicolumn{1}{c}{\makebox[0pt][c]{Train}} & \multicolumn{3}{c}{\makebox[0pt][c]{Test}} & Train Avg & Test Avg \\
\cmidrule(lr){2-2} \cmidrule(lr){3-5} 
& Obj1 & Obj2 & Obj3 & Obj4 &  & \\
\midrule
ACT (Vision-Only) & 5/5 & 0/5 & 4/5 & 0/5 & 100.0\% & 26.7\% \\
ACT (Vision+Force) & 5/5 & 3/5 & 4/5 & 4/5 & 100.0\% & 73.3\% \\
\midrule
FACTR & 5/5 & 4/5 & 5/5 & 5/5 & 100.0\% & 93.3\% \\
\bottomrule
\end{tabular}}
\caption{\small Comparison of methods for Fruit Pick-Place task.}
\label{tab:fruit_pick_place}
\end{table*}

\begin{table*}[t]
\centering
\resizebox{0.6\linewidth}{!}{%
\begin{tabular}{lcccccc}
\toprule
& \multicolumn{2}{c}{\makebox[0pt][c]{Train}} & \multicolumn{2}{c}{\makebox[0pt][c]{Test}} & Train Avg & Test Avg \\
\cmidrule(lr){2-3} \cmidrule(lr){4-5} 
& Obj1 & Obj2 & Obj3 & Obj4 &  & \\
\midrule
ACT (Vision-Only) & 0/5 & 0/5 & 0/5 & 0/5 & 0.0\% & 0.0\% \\
ACT (Vision+Force) & 4/5 & 4/5 & 3/5 & 4/5 & 80.0\% & 70.0\% \\
\midrule
FACTR & 5/5 & 4/5 & 4/5 & 4/5 & 90.0\% & 80.0\% \\
\bottomrule
\end{tabular}}
\caption{\small Comparison of methods for Rolling Dough task.}
\label{tab:rolling_dough}
\end{table*}




\end{document}